\documentclass[runningheads]{llncs}

\usepackage[T1]{fontenc}
\usepackage{graphicx}
\usepackage{amsmath,amssymb}
\usepackage[hidelinks]{hyperref}
\usepackage{subcaption}
\usepackage{xcolor}
\usepackage{microtype}
\newcommand\blfootnote[1]{%
  \begingroup\renewcommand\thefootnote{}\footnote{#1}\addtocounter{footnote}{-1}\endgroup}

\AtBeginDocument{\def\doi#1{}}

\begin{document}

\title{Macro-Action Topological Navigation under Noisy Localization using Reinforcement Learning}

\titlerunning{Macro-Action Topological Navigation under Noisy Localization}

\author{Simon Hakenes \and
Tobias Glasmachers}
\authorrunning{S. Hakenes and T. Glasmachers}
\institute{Institute of Neural Computation, Ruhr University Bochum, Germany\\
  \email{simon.hakenes@ini.rub.de, tobias.glasmachers@ini.rub.de}}

\maketitle
\blfootnote{Accepted at the Artificial Intelligence Symposium (AIS)~2026.}

\begin{abstract}
Navigating large, photorealistic 3D apartments from raw pixels is widely considered infeasible for plain reinforcement learning. We build an agent that does it anyway, estimating its own pose from the camera alone.
The agent has to reach several target objects in sequence, and their positions change between episodes, so it must explore to find them.
It builds on our earlier object-centric topological controller, which still read the agent's true pose and its object detections from the simulator.
Here we replace that true pose with an onboard, object-centric estimate.
For each object we keep a bank of ORB features that, when the object is seen again, yield a rough pose measurement, which a minimal Extended Kalman Filter (EKF) fuses with a motion model.
As on a real robot, the executed motions are noisy.
The estimate drifts, but the agent and the nearby objects drift together, so a locally consistent pose is enough to follow each short edge and then home in visually on the target, which lets us replace full SLAM with a much smaller model, closer to how biological navigation appears to work. 
In the photorealistic Habitat simulator, the agent reaches its target objects from vision alone, with a pose that only needs to be locally consistent.
\keywords{Deep Reinforcement Learning \and Navigation \and Localization \and Topological Maps \and Macro Actions}
\end{abstract}

\section{Introduction}
Consider how a person follows the instruction ``walk to the kitchen''.
They do not know the exact length of the corridor.
They only recall a few landmarks and roughly how these are arranged, and this is already enough to pick a sensible route and to keep track of where they roughly are while walking.
Localization here is local and approximate.
It only has to be good enough to decide where to turn next and to recognize the next landmark.
This is the intuition behind topological maps.
A topological map represents an environment as landmarks connected by traversable links, and not as a precise metric reconstruction.
Such a map does not need to be globally exact, or even globally consistent, to be useful for navigation.

There is evidence that mammals navigate in a similar way.
Animal cognitive maps seem to combine a loose topological structure, i.e., which places are connected, with only local metric information, rather than one globally consistent metric frame \cite{Alvernhe2012-tc,Poucet1993-mw}.
The study \cite{Parra-Barrero2023-xm} places spatial representations on a range from topological graphs to metric maps, and it points out that very different representations can produce the same behavior.
We take this as encouragement that a globally wrong but locally consistent map can be enough.

In our previous work \cite{Hakenes2025} we built a navigation agent around this approach.
We showed that a simple Deep Q-Network (DQN) \cite{Mnih2015-ut} agent can navigate large and realistic 3D scenes, which is normally considered infeasible for plain reinforcement learning (RL) from pixels.
The reason is that the agent does not act on raw pixels alone.
It operates on an object-centric topological map and selects macro actions, where one macro action means navigating to a known object.
The map turns a long and sparse-reward navigation episode into a short sequence of elementary decisions, which is what makes the RL problem tractable.
In other words, we give the agent a useful inductive prior in the form of structured world knowledge, such as objects and how to detect them, a graph, and the notion of moving through 3D space, instead of learning all of this from reward.

The task we study is sequential multi-object navigation.
In each episode the agent must reach several target objects in a fixed order, the same in every episode, which it has to learn.
Because the objects are placed differently each time, it cannot memorize a route and has to find them anew, building its topological map as it explores.
The closest benchmark is multi-object navigation (MultiON~\cite{Wani2020-km}); unlike MultiON, which reveals the next target class at each step and supplies a GPS-and-compass signal, our agent must learn which object each step denotes and localizes from vision alone.

However, that system quietly assumed perfect self-localization.
It read the true position and heading direction of the agent from the simulator and used them for localization, for placing objects on the map, and for executing macro actions.
This is convenient in a simulator but unrealistic for a real robot, where the pose has to be estimated from noisy sensors.
The classic answer to this problem is Simultaneous Localization and Mapping (SLAM), which builds a globally consistent metric map through probabilistic updates, loop closure, and bundle adjustment.
As argued above, however, brains do not seem to keep a globally consistent metric map, and our topological controller does not need one either.
If the agent only has to reach the next nearby waypoint, it is enough to know roughly where it is relative to that waypoint.
Avoiding global consistency is therefore both cheaper and closer to how biological navigation appears to work.

In this work we remove all ground-truth localization and test this idea directly.
We add a small object-centric localizer that estimates the pose from the camera alone.
For each object we keep a set of ORB image features \cite{Rublee2011-orb} together with their estimated 3D positions.
When the object is seen again, we match these features and align the matched points, which gives a rough measurement of position and heading.
A minimal Extended Kalman Filter (EKF) then fuses this measurement with a simple motion model of the agent.
The motion model provides yet another piece of world knowledge.
Like every roboticist, we roughly know what the agent does when it moves forward or turns, so we do not have to learn these dynamics from reward, even though the executed motion is noisy and the model is therefore only approximate.
The DQN is then left with the question of which object to visit, while the small model-based layer keeps track of where the agent is.

We run no loop closure and no global optimization, and each object is anchored once and then left in place, so the pose estimate drifts over time, but it drifts jointly with the nearby objects.
Because a macro action only follows short edges between nearby nodes, this locally consistent pose is enough for the controller.
A second reason is visual homing: as soon as the agent sees the target object, it can walk straight to it from the image alone, turning towards it in its visual array and stepping forward.
Visual homing is one of the basic navigation modes also found in animals, and it needs no accurate estimate of the global position \cite{FranzMallot2000,Parra-Barrero2023-xm,Trullier1997}.

\section{Related Work}

\paragraph{Classical SLAM.}
Estimating the agent's pose and a map of the environment at the same time is the subject of SLAM \cite{Dissanayake2001-dm}.
Feature-based visual systems such as ORB-SLAM \cite{Mur-Artal2015-cb} track local image features, often ORB \cite{Rublee2011-orb}, and build a globally consistent reconstruction through loop closure and bundle adjustment.
This machinery is what makes SLAM accurate, but it is more than our topological controller needs.
Bringing this much machinery to bear on a controller that only needs local consistency is like firing cannons at sparrows.
In our own trials it also proved fragile in simulation, where repetitive indoor textures and sparse features make tracking and loop closure unreliable.
We keep only the cheap front end, ORB features matched to per-object anchors and aligned with the Kabsch algorithm \cite{Kabsch1976-rt,Umeyama1991-ls}, and drop the global optimization.
That a locally consistent map can be enough for navigation has also been studied under the name of relative or topometric localization \cite{Mazuran2017-rt}, and earlier in topological SLAM that deliberately avoids a single global metric frame \cite{Choset2001-ck,Kuipers1991-dt}.

\paragraph{Localization in learned navigation.}
In embodied-AI benchmarks such as Habitat \cite{Savva2019-hh}, self-localization is often sidestepped by giving the agent a GPS-and-compass signal.
A common alternative is to estimate the pose from the visual stream, for example with a learned visual odometry module \cite{Zhao2021-vo}, or to learn a SLAM-like component that produces a metric map and a pose for the planner and policy, from RGB and odometry as in Active Neural SLAM \cite{Chaplot2020-pg}, or from RGB-D and semantics as in goal-oriented semantic exploration \cite{Chaplot2020-mb}.
These methods build a globally registered metric map with learned components.
We instead keep a topological map and a small classical localizer, and never optimize it into a globally consistent metric map.

\paragraph{Topological navigation.}
Topological maps are a natural representation for navigation, because they grow easily and make planning simple.
Semi-parametric topological memory \cite{Savinov2018-pg} stores observed views as graph nodes and learns a reachability network to localize against them without an explicit metric pose, and Neural Topological SLAM \cite{Singh_Chaplot2020-eg} learns to build and reason over a topological graph for image-goal navigation.
These systems learn their localization and graph reasoning, whereas we handle localization with a small classical stack and use the DQN only for the high-level choice of which object to visit.

\paragraph{Macro actions.} 
Selecting a high-level target that unfolds into a long sequence of elementary actions is a form of temporal abstraction, formalized by the options framework \cite{Sutton1999-yp}.
Unlike learned options, our macro actions are not trained from reward but built from the map and a small controller, so the RL problem reduces to choosing useful objects, scored by one shared network over the growing action set following the DRRN \cite{He2016-ui}.
This design is from our previous system \cite{Hakenes2025}, and we ask whether it still holds once the true pose becomes a noisy, locally consistent estimate.

\section{Method}
\label{sec:method}

Before going into detail, we give a high-level overview of the complete system and clarify which parts are inherited from our previous work \cite{Hakenes2025} and which are new in this paper.

\begin{figure}[t]
    \centering
    \includegraphics[width=\textwidth]{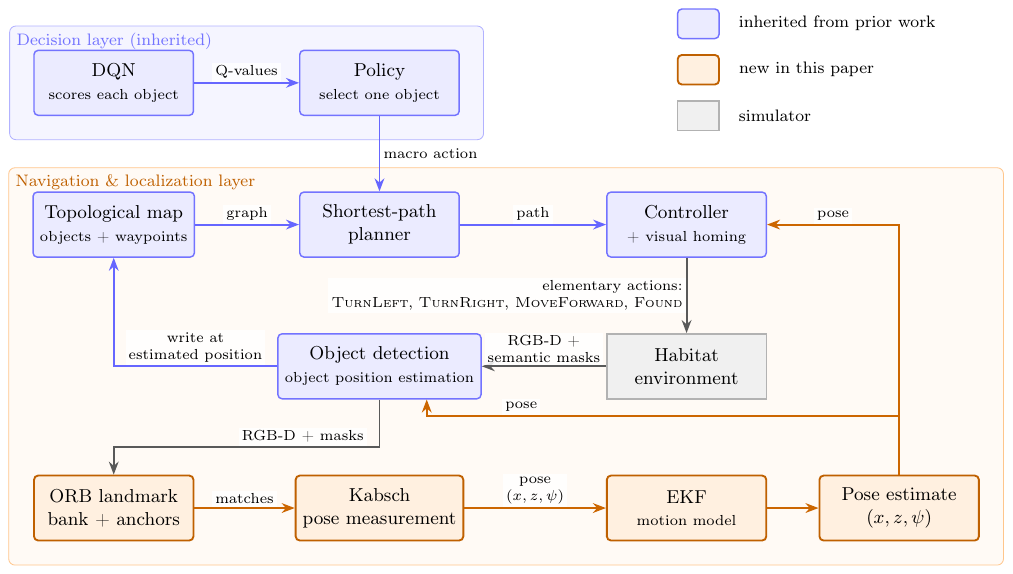}
    \caption{Overview of the system.
    The decision layer (DQN over object-centric macro actions, blue) is inherited from our previous work \cite{Hakenes2025}.
    The navigation and localization layer turns the selected object into motion and replaces the previous ground-truth pose oracle with the new localization stack (orange).
    An ORB landmark bank, a pose measurement, and an EKF with a motion model produce a pose estimate.
    It feeds the controller, and it is used to estimate the positions of detected objects, which are written to the topological map.
    Close to the target the controller switches to visual homing.}
    \label{fig:overview}
\end{figure}

The method is organized into two layers (Fig.~\ref{fig:overview}).
The first one is the decision layer, which is the object-centric macro-action DQN from our previous work \cite{Hakenes2025}.
It takes the current task progress and the objects stored in the map, and decides which known object should be visited next.
The second one is the navigation and localization layer.
It turns that decision into actual motion, plans a shortest path over the topological map with a standard shortest-path algorithm, and drives the agent there with a small set of elementary actions (moving forward and turning).
This split between the two layers is the design we keep from our previous work.
Sparse-reward deep RL should not have to rediscover 3D geometry, graph search, or the effect of a forward step from raw pixels.
We make these parts explicit, and we leave the DQN with the semantic problem of choosing useful objects to interact with.

What is new is the localization stack that lets this layer work without the true pose.
In our previous system this layer could read the true simulator pose at every step, so localization and map construction were trivial.
Here we replace this oracle with a localizer that estimates the pose from the agent's own sensors.
For each object we keep a bank of ORB landmarks, we turn re-encountered landmarks into a rough pose measurement by aligning them, and an EKF with a motion model fuses these measurements with the known geometry of the elementary actions.
The motion model alone is not enough, because a forward step can be blocked by an obstacle, the agent can slide along a wall, or the actuation itself is noisy, so the predicted motion and the real motion drift apart.
Because the pose is now an estimate and not ground-truth information, we also add a map-write gate that refuses to write new map entries when the pose estimate is too unreliable.

\subsection{Macro actions and the state-action space}
\label{sec:method_state_action}

The agent builds and uses an object-centric topological map, both in our previous system and here.
A macro action means navigating to a specific object that is already stored in this map, which turns a long and sparse-reward navigation episode into a shorter sequence of semantic choices.
Each macro action runs until the agent performs a \textsc{Found} action, which declares that it has reached the chosen object and ends the macro.

The state and action spaces are not fixed, because the map grows while the episode runs.
Let $\mathcal{O}_t \subset \mathcal{V}_t$ be the object nodes that are known at time $t$.
The macro-action space is the set of these objects, $A_t = \mathcal{O}_t$.

For an object node $v \in \mathcal{O}_t$, the map stores a set of image patches
\begin{equation}
  \mathcal{F}_v = \{f_{v,1}, \ldots, f_{v,n_v}\},
  \qquad
  f_{v,i} \in \mathbb{R}^{H \times W \times 3}.
  \label{eq:features}
\end{equation}
The task progress is encoded by a one-hot vector $g_t \in \{0,1\}^{N_T}$, where $N_T$ is the number of targets in the sequence.
The active entry indicates which step in the fixed sequence is currently active, and the agent has to learn which object that step refers to.

We keep this representation from the previous paper, because it solves the main difficulty of using a growing map with a DQN.
A standard DQN has one output neuron per action \cite{Mnih2015-ut}, which assumes a fixed set of actions.
Here the number of objects is not known in advance, so the network cannot output all action values at once.
Instead, inspired by the DRRN \cite{He2016-ui}, it evaluates one object at a time, $Q_\theta(\mathcal{F}_v, g_t) \in \mathbb{R}$, where the object is an input to the network.
The same network is applied to every known object and produces one value per candidate macro action.

\subsection{DQN architecture and policy}
\label{sec:method_dqn}

For each object we use up to $N_i$ image patches as the action representation for robustness.
All patches pass through the same convolutional feature extractor.
We then combine the resulting feature vector with the progress vector through an outer product.
A fully connected layer maps the combined features to a single Q-value.

At decision time the network computes one Q-value for each known object.
The policy is greedy most of the time, and it explores with a Boltzmann policy \cite{Wiering1999-sv} that samples an object from a softmax over the Q-values.
Training follows the same object-wise evaluation as in~\cite{Hakenes2025}.

\subsection{Map representation}
\label{sec:method_topomap}

The map at time $t$ is an undirected weighted graph $\mathcal{G}_t = (\mathcal{V}_t, \mathcal{E}_t)$.
The node set $\mathcal{V}_t = \mathcal{B}_t \cup \mathcal{O}_t$ contains backbone nodes $\mathcal{B}_t$ and object nodes $\mathcal{O}_t$.
A backbone node is simply a pose that the agent has already visited.
It has no visual appearance and only serves as a waypoint for navigation.
An object node represents a semantic object and stores an object id, an estimated 3D position, and a set of image patches.
Only object nodes are offered to the DQN as macro actions.
The backbone nodes are internal navigation scaffolding.

This distinction matters.
An object node is a good semantic landmark but a poor waypoint, because an object's position can lie inside furniture, behind an occluder, or be visible but not reachable.
A backbone node, in contrast, is a pose that the agent has actually occupied, so it is usually safe to navigate to.
When the DQN selects an object, the controller therefore first routes to a nearby backbone node and only switches to visual homing once the object comes into view.

It is important to keep in mind that the map is not given in advance.
At the start of an episode the map is empty, and the agent has usually not even seen the target objects yet, because they lie elsewhere in an apartment that typically spans several rooms.
The agent first has to explore the environment and build the map, and it already navigates on this partial map while the map is still growing.
After each elementary action, the current estimated pose is added as a backbone node, and nearby backbone nodes are connected.

Object nodes are added from the RGB-D image and the semantic masks.
As in the previous paper, the masks and object ids come from the simulator.
Object detection is a separate problem and not the topic of this paper, so we do not address it here.
For every visible object, we store the RGB pixels inside its bounding box as a patch, and we back-project the depth at the object to get a 3D position in the map frame.
If the object is new, a new object node is created.
If it is already known, we may add a better (bigger) patch, but we do not move the stored object position afterwards.
The map is therefore not globally consistent.
One could object that a single global coordinate frame is a metric map through the back door.
However, nothing in the controller depends on absolute coordinates.
Because we never enforce global consistency, the absolute coordinates drift, but what the agent actually relies on is the relative, node-to-node geometry, which is why the global pose can drift by~meters without breaking navigation.

\subsection{ORB object landmarks}
\label{sec:method_orb}

The semantic masks tell us which object is visible, but they do not give stable point correspondences over time, because a mask only labels which pixels belong to an object and does not identify the same physical point from one frame to the next.
For localization we therefore keep a small bank of ORB features \cite{Rublee2011-orb} for each object.
We use ORB for practical reasons.
The features are cheap compared with learned descriptors, they can be matched quickly, and they need no training.

For every visible object, ORB keypoints are detected only inside its semantic mask, so each feature belongs to exactly one object.
When a feature is seen for the first time, we store its descriptor together with a 3D anchor, which we obtain by back-projecting its depth into the map frame.
When the same object is seen again, we match the current descriptors against the object's bank.
Each successful match links a freshly observed point to its stored anchor on the map.
These matched pairs are the input to the pose measurement described next.

Only re-found anchors are used for localization.
Newly detected keypoints can extend the bank of an object for later, but they are not used as localization correspondences in the frame in which they first appear.
This avoids a circular update in which a newly seen object both defines and confirms the current pose.

\subsection{Kabsch pose measurement}
\label{sec:method_kabsch}

The localizer turns the matched ORB correspondences into a single noisy pose measurement.
Each matched feature gives two points on the ground plane.
One is where the agent currently observes it, obtained by back-projecting its depth.
The other is where its anchor is stored on the map.
Up to the unknown pose of the agent, these two point sets should be connected by a rigid transformation.
Since the agent moves on the ground plane, this motion is a 2D translation together with a single rotation around the vertical axis, i.e., an SE(2) change in position and yaw.
Given at least three matches, we recover this motion with the Kabsch algorithm \cite{Kabsch1976-rt,Umeyama1991-ls}.

Formally, let $(p_i, q_i)$ with $i = 1, \ldots, M$ be the matched pairs on the ground plane, where $p_i$ is the back-projected observation in the agent frame and $q_i$ is the stored anchor in map coordinates.
Each pair carries an inverse-distance weight $w_i$ with $\sum_i w_i = 1$, so that far and therefore noisier anchors contribute less.
With the weighted centroids $\bar{p} = \sum_i w_i\, p_i$ and $\bar{q} = \sum_i w_i\, q_i$, the rotation follows from the SVD of the weighted cross-covariance
\begin{equation}
  C = \sum_{i} w_i\, (p_i - \bar{p})(q_i - \bar{q})^\top = U S V^\top
  \label{eq:kabsch-cov}
\end{equation}
as
\begin{equation}
  R = V \operatorname{diag}\!\left(1, \det(V U^\top)\right) U^\top,
  \qquad
  \mathbf{t} = \bar{q} - R\, \bar{p}.
  \label{eq:kabsch-rt}
\end{equation}
The determinant factor forces a proper rotation rather than a reflection, since the agent can turn but not mirror itself.
The agent sits at the origin of its own frame, so $\mathbf{t}$ is directly the measured position, and the rotation angle of $R$ is the measured yaw.
Together they form one measurement of position and heading that is handed to the EKF.

The measurement only has to pull the predicted pose back towards the already mapped object anchors from time to time.
The alignment also returns the residuals $r_i = \lVert R\, p_i + \mathbf{t} - q_i \rVert$, which tell how well the matches agree, and we keep their distribution as a measure of how much we can trust this measurement.

\subsection{Extended Kalman filter}
\label{sec:method_ekf}

The EKF fuses the geometry of the elementary actions with the Kabsch measurements into a single pose estimate that is updated at every step.
The motion model behind its prediction step is the piece of world knowledge mentioned in the introduction.

The pose estimate is represented by the SE(2) state
\begin{equation}
  \mathbf{s}_t =
  \begin{bmatrix}
  x_t & z_t & \psi_t
  \end{bmatrix}^{\top},
  \label{eq:state}
\end{equation}
where $(x,z)$ are ground-plane coordinates and $\psi$ is yaw.
The prediction step is tied to the commanded elementary action.
For a forward step of length $d$,
\begin{equation}
  x_{t+1} = x_t - d\sin\psi_t,\qquad
  z_{t+1} = z_t - d\cos\psi_t,\qquad
  \psi_{t+1} = \psi_t .
  \label{eq:motion}
\end{equation}
For elementary \textsc{TurnLeft} and \textsc{TurnRight} actions of the low-level controller, the position stays the same and the yaw changes by the known turn angle.

We deliberately add actuation noise, so that the motion model is no longer exact, especially in the yaw.
The executed motion deviates from the commanded amounts by zero-mean Gaussian noise in three components, with standard deviations of $1^\circ$ on the turn angle, $0.01$~m on the forward step, and $0.005$~m of sideways slip, as on a real robot.
The EKF still predicts the commanded motion, so the believed pose drifts away from the true pose like real odometry.

This gives the agent a pose estimate even in frames where no object can be matched, for example when it stands close to a blank wall and sees no useful features.

The Kabsch output enters as a direct measurement whose assumed noise grows as the matches agree less, and measurements that are clearly inconsistent with the current belief are rejected.
The filter therefore coasts on the motion model through unreliable visual frames and accepts a visual correction only when it fits the current estimate.

\subsection{Map-write gating}
\label{sec:method_map_write_gate}

A noisy pose creates a failure mode that did not exist when the true pose was available: if the pose is wrong, an object can be written to the map at a wrong position, and this wrong landmark then misleads all later planning and navigation.
We therefore only write to the map when the current pose looks trustworthy, judged from simple signals the localizer already provides, such as how many object features were matched and how certain the EKF currently is.
If the pose is not good enough, we set the observation aside and add the object later, since we would rather add an object late than in the wrong place.

\subsection{Macro-action execution under noisy localization}
\label{sec:method_controller}

To execute a macro action, the controller turns the chosen object into a navigation target.
It then plans the shortest path through the graph and follows it one waypoint at a time, turning towards the next waypoint and moving forward.
The controller is deliberately simple, though it still needs fallback behaviors to stay reliable.
Its job is not to solve navigation in arbitrary geometry, but to use the fact that the graph splits a long route into short and tractable steps.

Close to the target object, the controller switches to visual homing.
The agent reads the position of the target in the input image from the semantic mask and the depth.
If the target is left or right of the image center, the agent turns towards it.
If it is centered, the agent moves forward.
If it is close enough, the controller signals \textsc{Found}, which ends the macro action.
A fallback also ends the macro if the agent stalls without reaching the object, so it cannot run forever.
Visual homing is reliable here, because close to the object the camera image tells the agent directly where to go, and it does not need an accurate estimate of its global position.
This is the main reason why a drifting global pose does not hurt.
The drift matters least exactly where the agent has to be accurate, namely right at the target.
Visual homing relies on the live camera image rather than the believed pose, so it ties the agent to the real object however far the global estimate has drifted, and seeing that object again lets the localizer pull the estimate back towards its anchor.

\section{Experiments and Results}
\label{sec:experiments}

We evaluate in the photorealistic Habitat simulator \cite{Savva2019-hh} on the same multi-room object-finding task as in our previous work \cite{Hakenes2025}.
The agent starts in an unknown apartment with an empty map and must reach a fixed sequence of target objects in the correct order.
Because it does not know where the targets are, it has to explore, build the topological map online, and revisit objects it has already seen, and an episode counts as a success only when every target in the sequence has been reached. 
We use two reward schemes: under \emph{immediate} reward the agent is rewarded each time it reaches the next target in the order, whereas under \emph{terminal} reward it is rewarded only once, when the whole sequence is complete.
Terminal reward is the harder credit-assignment problem, because a long chain of correct macro actions is reinforced only at the very end, and with a single target the two schemes coincide.
As before, object masks and ids are provided by the simulator, so the new element under test is purely the localization stack, not object detection.
The central question is whether the macro-action navigation that previously relied on ground-truth pose still works when that pose is replaced by our noisy, locally consistent estimate.

We report the success rate and the number of elementary actions per episode, and Table~\ref{tab:results} additionally lists the localization error as per-episode median RMSE against the ground-truth pose.
Each condition and task type is trained with five independent seeds.

To see what each part contributes, we compare five conditions.
Full is the complete system with the ORB, Kabsch, and EKF localizer.
No-EKF is an ablation that uses the raw Kabsch measurement directly, without the motion model and the filter.
Motion-Only is the complementary ablation that keeps the EKF motion model but drops the Kabsch measurement, so the pose comes from dead reckoning alone.
GT-Pose is an oracle that restores the true pose and, unlike the other four conditions, runs without actuation noise, reproducing our previous system \cite{Hakenes2025}.
Random-Macro keeps the full localizer but picks the macro actions at random instead of using the trained DQN.

\begin{figure}
    \centering
    \includegraphics[width=\textwidth]{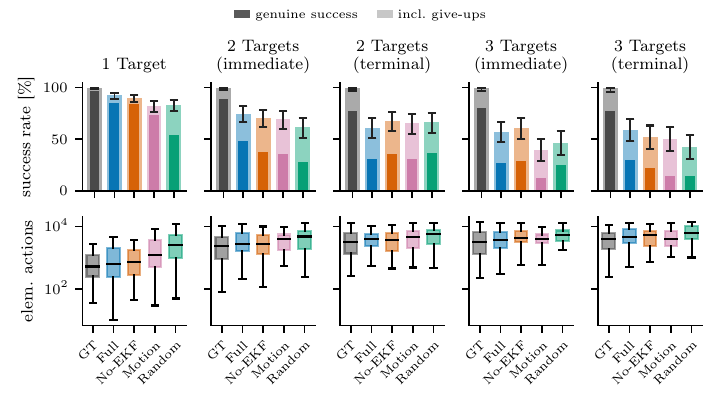}
    \caption{Success rate (top, with 95\% confidence intervals) and elem.\ actions per successful episode (bottom, note the logarithmic scale) for the five conditions and the five task types.
    The success bars are two-toned: the light bar is the conventional success rate, and the solid inner bar is the genuine success rate, so the light remainder is the share won by give-ups, i.e., a \textsc{Found}, declared after the agent stalls without reaching the target, that happens to land inside the success radius.
    Failed episodes are excluded from the action counts because they end at the step budget, so their length reflects the budget and not the navigation.}
    \label{fig:results}
\end{figure}

\begin{figure}[t]
    \centering
    \begin{subfigure}[b]{0.215\textwidth}\centering
        \includegraphics[width=\linewidth]{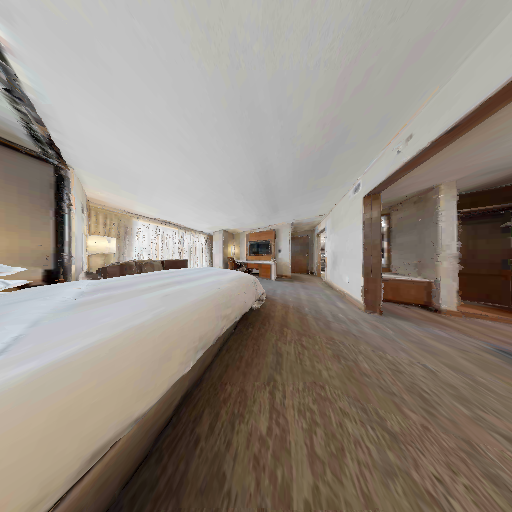}
        \caption{onboard view}
        \label{fig:metrics-rgb}
    \end{subfigure}\hfill
    \begin{subfigure}[b]{0.31\textwidth}\centering
        \includegraphics[width=\linewidth]{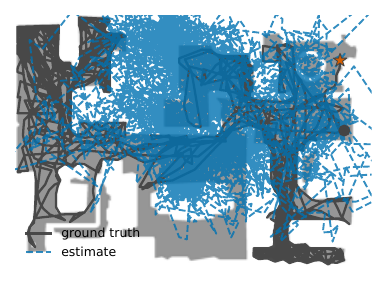}
        \caption{map + path}
        \label{fig:metrics-topdown}
    \end{subfigure}\hfill
    \begin{subfigure}[b]{0.235\textwidth}\centering
        \includegraphics[width=\linewidth]{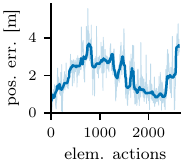}
        \caption{position drift}
        \label{fig:metrics-pos}
    \end{subfigure}\hfill
    \begin{subfigure}[b]{0.235\textwidth}\centering
        \includegraphics[width=\linewidth]{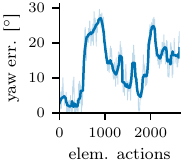}
        \caption{heading error}
        \label{fig:metrics-yaw}
    \end{subfigure}
    \caption{A single successful episode of the Full system, with the onboard view (a), a bird's-eye map showing the ground-truth (solid) and estimated (dashed) paths (b), and the position (c) and yaw (d) error over the episode.
    The estimated path in (b) drifts visibly away from the ground truth, so the global pose really is inconsistent.
    Panels (c) and (d) show the position error growing to a few~meters and the heading error to roughly fifteen to thirty~degrees over the episode, yet the controller still reaches each target, because the agent and the nearby objects drift together, so only their relative geometry matters.}
    \label{fig:metrics}
\end{figure}

Table~\ref{tab:results} summarizes the five conditions, Fig.~\ref{fig:results} shows success rates and navigation cost per condition, and Fig.~\ref{fig:metrics} shows the behavior of the localizer over one episode.
We first look at task success under the conventional criterion, where an episode counts as a success whenever the agent declares a target \textsc{Found} within a small radius of it.
By this measure the five conditions stay close together.
Even Random-Macro, which picks its macro actions without any trained policy, reaches 84\%, 62\%, and 46\% of the targets at one, two, and three targets.
This is surprising, because the localization errors in Table~\ref{tab:results} differ widely across the conditions, and one would expect a broken pose to hurt navigation.

The \textsc{Found} declarations explain the discrepancy.
A wrong \textsc{Found} carries no penalty, so an agent that never gets to its target can give up after searching for a while and declare \textsc{Found} anyway, and this declaration sometimes lands inside the success radius by chance.
The conventional rate counts these give-ups together with the deliberate arrivals.
We therefore also report the genuine success rate, which counts only the episodes in which the agent actually reached the target and stopped at it.
Under this measure the conditions separate, most clearly in the single-target and immediate-reward cells where every target is rewarded.
The Full system loses little at a single target, from 93\% to 85\%, and although the gap grows with the sequence length, its declared successes remain largely genuine arrivals rather than give-ups.
Random-Macro loses far more at every length, from 84\% to 54\%, from 62\% to 28\%, and from 46\% to 25\%.
The gap is starkest for Random-Macro: at a single target its 84\% conventional rate falls to 54\% genuine, the give-ups making up the difference.
The gap between the two rates is the share of successes won by chance, and it is largest for the random baseline.
Under terminal reward, where only the completed sequence is rewarded, the genuine rates are low and close across the noisy conditions.
Three targets under terminal reward is the hardest cell, since the whole sequence must be completed before any reward arrives; even here the Full system completes most sequences and keeps the highest genuine success of the noisy conditions.

The localization error separates the conditions more sharply still.
Full keeps its heading error between 13 and 39~degrees and its position drift to a few~meters across all task types.
No-EKF lets the heading error grow to between 80 and 104~degrees, so its heading estimate is essentially unusable.
Motion-Only drifts in position to between 3 and 4~meters.
We return to both ablations in the discussion.
The elementary-action counts must be read together with the success rate, since they are taken over successful episodes only.
A condition that fails often, e.g., Motion-Only at three targets, can still look cheap, because the few episodes it wins are the short, easy ones.
GT-Pose, the oracle, reaches every target and keeps the highest genuine success of all, so a noisy pose does cost some performance, but among the conditions that must work from a noisy pose the Full system is the strongest.

\begin{table}[!t]
    \centering
    \caption{Comparison of the five conditions per task type, pooled over five training seeds. $n$ is the number of evaluation episodes. Pos.\ and Yaw RMSE are per-episode medians taken over all episodes. Lower step counts and errors are better.}
    \label{tab:results}
    \setlength{\tabcolsep}{4pt}
    \small
    \begin{tabular}{lcccccc}
        \hline
        Condition & Success & Genuine & Elem.\ actions & Pos.\ RMSE & Yaw RMSE & $n$ \\
                  &         & succ.\  & [median]       & [m]        & [deg]    &     \\
        \hline
        \multicolumn{7}{l}{\textit{1 Target}} \\
        GT-Pose (oracle) & 100.0\% & 96.9\% & 530 & 0.00 & 0.0 & 426 \\
        Full (ours) & 92.7\% & 85.4\% & 646 & 1.46 & 13.2 & 314 \\
        No-EKF & 90.3\% & 84.0\% & 737 & 3.80 & 79.7 & 300 \\
        Motion-Only & 82.6\% & 73.4\% & 1257 & 2.67 & 18.4 & 184 \\
        Random-Macro & 83.5\% & 54.1\% & 2643 & 2.01 & 27.6 & 170 \\
        \hline
        \multicolumn{7}{l}{\textit{2 Targets, immediate reward}} \\
        GT-Pose (oracle) & 100.0\% & 89.2\% & 2411 & 0.00 & 0.0 & 158 \\
        Full (ours) & 74.8\% & 48.7\% & 2818 & 2.08 & 26.5 & 119 \\
        No-EKF & 70.9\% & 37.6\% & 2804 & 5.75 & 101.6 & 117 \\
        Motion-Only & 69.2\% & 35.5\% & 4121 & 3.71 & 33.4 & 107 \\
        Random-Macro & 61.5\% & 28.1\% & 4796 & 2.11 & 40.0 & 96 \\
        \hline
        \multicolumn{7}{l}{\textit{2 Targets, terminal reward}} \\
        GT-Pose (oracle) & 100.0\% & 76.9\% & 3207 & 0.00 & 0.0 & 130 \\
        Full (ours) & 61.3\% & 31.2\% & 3912 & 2.20 & 37.7 & 93 \\
        No-EKF & 68.0\% & 36.0\% & 3744 & 5.61 & 97.5 & 100 \\
        Motion-Only & 65.6\% & 31.1\% & 4503 & 3.69 & 35.0 & 90 \\
        Random-Macro & 66.7\% & 36.8\% & 5866 & 2.19 & 40.2 & 87 \\
        \hline
        \multicolumn{7}{l}{\textit{3 Targets, immediate reward}} \\
        GT-Pose (oracle) & 100.0\% & 79.8\% & 3125 & 0.00 & 0.0 & 109 \\
        Full (ours) & 57.3\% & 27.0\% & 3811 & 2.26 & 36.0 & 89 \\
        No-EKF & 61.0\% & 29.3\% & 4445 & 6.29 & 104.2 & 82 \\
        Motion-Only & 39.2\% & 12.2\% & 4058 & 3.95 & 43.3 & 74 \\
        Random-Macro & 46.4\% & 24.6\% & 5260 & 2.22 & 44.8 & 69 \\
        \hline
        \multicolumn{7}{l}{\textit{3 Targets, terminal reward}} \\
        GT-Pose (oracle) & 100.0\% & 77.0\% & 3911 & 0.00 & 0.0 & 87 \\
        Full (ours) & 59.2\% & 30.3\% & 4693 & 2.36 & 39.2 & 76 \\
        No-EKF & 52.1\% & 22.5\% & 5402 & 6.50 & 98.7 & 71 \\
        Motion-Only & 50.0\% & 14.7\% & 4028 & 4.39 & 34.7 & 68 \\
        Random-Macro & 42.2\% & 14.1\% & 6393 & 2.46 & 48.6 & 64 \\
        \hline
    \end{tabular}
\end{table}

\section{Discussion and Conclusion}
\label{sec:discussion}

Our guiding question was whether macro-action topological navigation still works once the global pose is noisy, i.e., whether a pose that is only locally consistent is sufficient.
The experiments support a positive answer, with one qualification.
The Full system reaches its targets even though its global position estimate drifts by several~meters, and on the stricter genuine measure it leads the conditions that work from a noisy pose.
It does not match the ground-truth oracle, and the gap widens as the sequence grows longer.
We therefore do not claim that the noise is harmless, only that local consistency is enough for the controller to keep working.

The No-EKF ablation shows which part of the localizer matters.
Without the motion model the heading estimate drifts to between 80 and 104~degrees and navigation breaks down.
Hence it is the explicit motion model, and not the raw Kabsch measurement, that makes the pose usable.
The low action counts of No-EKF are the selection effect from the results, since the episodes it wins are the easy ones and its failures show up in the success rate instead.

The Motion-Only ablation is the complementary cut, and it is where the actuation noise matters most.
It keeps the motion model but drops the Kabsch measurement, so the pose is pure dead reckoning.
Because the executed motions are noisy, as they would be on a real robot, this drift is never corrected, and the position error grows to several~meters while the heading error reaches 18 to 43~degrees.
In our earlier noise-free setup this condition looked almost perfect, which was an artifact of the simulator applying every commanded move exactly.
With realistic actuation noise that artifact is gone, and Motion-Only now shows the unbounded drift that the Kabsch measurement is there to correct.
The two ablations together separate the two jobs of the localizer.
The motion model keeps the heading usable between measurements, and the measurement keeps the position from drifting away.

A natural question is whether the system needs the depth image at all.
Depth enters in only two places: it back-projects the ORB matches and the detected objects into 3D anchors, and it tells visual homing when the target is close enough.
Neither use needs a dense or globally accurate depth map.
Because we only ever require local consistency, the metric scale that depth provides could in principle come from cheaper sources, such as triangulation of the ORB anchors across the known elementary motions, a learned monocular depth estimate, or even the apparent size of an object's mask.
The same tolerance to drift that lets us drop global SLAM should make the localizer forgiving of such a monocular substitute, which we see as a promising route to removing the depth sensor.

Three limitations stand out.
First, computing ORB features inside every object mask is the most expensive part of the localizer, so the localization stack is comparatively costly, and cheaper or learned features are an obvious target for future work.
Second, we still rely on the simulator for the object masks and ids, so we do not yet deal with real object detection.
Adding real object detection, together with the noise it would inject into the per-object feature banks, is an important step towards a fully realistic system.
Third, the macro actions are carried out by a hand-built controller that follows waypoints, homes in visually, and falls back to recovery behaviors when the agent stalls.
Much of the navigation reliability, and most of the step count, rests on this controller rather than on the learned policy, which is also why even the Random-Macro baseline reaches its targets fairly often.
Making such a controller robust across diverse and cluttered apartments takes a good deal of manual engineering, so learning more of this low-level execution is an appealing direction for future work.

In sum, removing the ground-truth pose oracle and navigating from a globally inaccurate but locally consistent estimate works well, though not yet as reliably as with the true pose.
The cost of the ORB features and the assumption of perfect object detection are the main steps left before the system can run outside the simulator.

\bibliographystyle{splncs04}
\bibliography{bibliography}

@ARTICLE{Dissanayake2001-dm,
  title     = "{A solution to the simultaneous localization and map building
               (SLAM) problem}",
  author    = "Dissanayake, M W M G and Newman, P and Clark, S and
               Durrant-Whyte, H F and Csorba, M",
  journal   = "IEEE Trans. Rob. Autom.",
  publisher = "Institute of Electrical and Electronics Engineers (IEEE)",
  volume    =  17,
  number    =  3,
  pages     = "229--241",
  month     =  jun,
  year      =  2001,
  doi       = "10.1109/70.938381",
  issn      = "1042-296X,2374-958X"
}

@PHDTHESIS{Wiering1999-sv,
  title  = "{Explorations in efficient reinforcement learning}",
  author = "Wiering, Marco",
  year   =  1999,
  school = "University of Amsterdam"
}

@InProceedings{Hakenes2025,
author="Hakenes, Simon
and Glasmachers, Tobias",
editor="Nicosia, Giuseppe
and Ojha, Varun
and Giesselbach, Sven
and Pardalos, Panos M.
and Umeton, Renato
and La Malfa, Emanuele
and La Malfa, Gabriele",
title="Deep Reinforcement Learning Based Navigation with Macro Actions and Topological Maps",
booktitle="Machine Learning, Optimization, and Data Science",
year="2026",
publisher="Springer Nature Switzerland",
address="Cham",
pages="95--108",
isbn="978-3-032-21477-5",
doi="10.1007/978-3-032-21477-5_7"
}

@ARTICLE{Sutton1999-yp,
  title   = "{Between MDPs and semi-MDPs: A framework for temporal abstraction
             in reinforcement learning}",
  author  = "Sutton, Richard S and Precup, Doina and Singh, Satinder",
  journal = "Artif. Intell.",
  volume  =  112,
  number  = "1--2",
  pages   = "181--211",
  month   =  aug,
  year    =  1999,
  doi     = "10.1016/S0004-3702(99)00052-1",
  issn    = "0004-3702"
}

@INPROCEEDINGS{Chaplot2020-pg,
  title     = "{Learning to Explore using Active Neural {SLAM}}",
  author    = "Chaplot, Devendra Singh and Gandhi, Dhiraj and Gupta, Saurabh and Gupta, Abhinav and Salakhutdinov, Ruslan",
  booktitle = "{International Conference on Learning Representations (ICLR)}",
  year      =  2020
}

@INPROCEEDINGS{Singh_Chaplot2020-eg,
  title     = "{Neural topological SLAM for visual navigation}",
  author    = "Chaplot, Devendra Singh and Salakhutdinov, Ruslan and Gupta,
               Abhinav and Gupta, Saurabh",
  booktitle = "{2020 IEEE/CVF Conference on Computer Vision and Pattern
               Recognition (CVPR)}",
  publisher = "IEEE",
  pages     = "12875--12884",
  month     =  jun,
  year      =  2020,
  doi       = "10.1109/cvpr42600.2020.01289",
  isbn      =  9781728171685
}

@ARTICLE{Chaplot2020-mb,
  title   = "{Object goal navigation using goal-oriented semantic exploration}",
  author  = "Chaplot, Devendra Singh and Gandhi, Dhiraj Prakashchand and Gupta,
             Abhinav and Salakhutdinov, Russ R",
  journal = "Adv. Neural Inf. Process. Syst.",
  volume  =  33,
  pages   = "4247--4258",
  year    =  2020,
  issn    = "1049-5258"
}

@ARTICLE{Alvernhe2012-tc,
  title   = "{Rats build and update topological representations through
             exploration}",
  author  = "Alvernhe, Alice and Sargolini, Francesca and Poucet, Bruno",
  journal = "Anim. Cogn.",
  volume  =  15,
  number  =  3,
  pages   = "359--368",
  month   =  may,
  year    =  2012,
  doi     = "10.1007/s10071-011-0460-z",
  issn    = "1435-9448,1435-9456"
}

@ARTICLE{Parra-Barrero2023-xm,
  title    = "{A map of spatial navigation for neuroscience}",
  author   = "Parra-Barrero, Eloy and Vijayabaskaran, Sandhiya and Seabrook,
              Eddie and Wiskott, Laurenz and Cheng, Sen",
  journal  = "Neurosci. Biobehav. Rev.",
  volume   =  152,
  pages    =  105200,
  month    =  sep,
  year     =  2023,
  doi      = "10.1016/j.neubiorev.2023.105200",
  pmid     =  37178943,
  issn     = "0149-7634,1873-7528",
  language = "en"
}

@INPROCEEDINGS{Wani2020-km,
  title     = "{MultiON: Benchmarking Semantic Map Memory using Multi-Object
               Navigation}",
  author    = "Wani, Saim and Patel, Shivansh and Jain, Unnat and Chang, Angel
               and Savva, Manolis",
  editor    = "Larochelle, H and Ranzato, M and Hadsell, R and Balcan, M F and
               Lin, H",
  booktitle = "{Advances in Neural Information Processing Systems}",
  publisher = "Curran Associates, Inc.",
  volume    =  33,
  pages     = "9700--9712",
  year      =  2020
}

@ARTICLE{Mur-Artal2015-cb,
  title   = "{ORB-SLAM: A Versatile and Accurate Monocular SLAM System}",
  author  = "Mur-Artal, Ra\'{u}l and Montiel, J M M and Tard\'{o}s, Juan D",
  journal = "IEEE Trans. Rob.",
  volume  =  31,
  number  =  5,
  pages   = "1147--1163",
  month   =  oct,
  year    =  2015,
  doi     = "10.1109/TRO.2015.2463671",
  issn    = "1941-0468"
}

@INPROCEEDINGS{Savva2019-hh,
  title     = "{Habitat: A platform for embodied AI research}",
  author    = "Savva, Manolis and Kadian, Abhishek and Maksymets, Oleksandr and Zhao, Yili and Wijmans, Erik and Jain, Bhavana and Straub, Julian and Liu, Jia and Koltun, Vladlen and Malik, Jitendra and Parikh, Devi and Batra, Dhruv",
  booktitle = "{2019 IEEE/CVF International Conference on Computer Vision
               (ICCV)}",
  pages     = "9338--9346",
  publisher = "IEEE",
  month     =  oct,
  year      =  2019,
  doi       = "10.1109/iccv.2019.00943",
  isbn      =  9781728148038
}

@ARTICLE{Mnih2015-ut,
  title    = "{Human-level control through deep reinforcement learning}",
  author   = "Mnih, Volodymyr and Kavukcuoglu, Koray and Silver, David and Rusu,
              Andrei A and Veness, Joel and Bellemare, Marc G and Graves, Alex
              and Riedmiller, Martin and Fidjeland, Andreas K and Ostrovski,
              Georg and Petersen, Stig and Beattie, Charles and Sadik, Amir and
              Antonoglou, Ioannis and King, Helen and Kumaran, Dharshan and
              Wierstra, Daan and Legg, Shane and Hassabis, Demis",
  journal  = "Nature",
  volume   =  518,
  number   =  7540,
  pages    = "529--533",
  month    =  feb,
  year     =  2015,
  doi      = "10.1038/nature14236",
  pmid     =  25719670,
  issn     = "0028-0836,1476-4687",
  language = "en"
}

@ARTICLE{Poucet1993-mw,
  title    = "{Spatial cognitive maps in animals: new hypotheses on their
              structure and neural mechanisms}",
  author   = "Poucet, Bruno",
  journal  = "Psychol. Rev.",
  volume   =  100,
  number   =  2,
  pages    = "163--182",
  month    =  apr,
  year     =  1993,
  doi      = "10.1037/0033-295x.100.2.163",
  pmid     =  8483980,
  issn     = "0033-295X",
  language = "en"
}

@INPROCEEDINGS{He2016-ui,
  title     = "{Deep Reinforcement Learning with a Natural Language Action Space}",
  author    = "He, Ji and Chen, Jianshu and He, Xiaodong and Gao, Jianfeng and Li,
               Lihong and Deng, Li and Ostendorf, Mari",
  booktitle = "Proceedings of the 54th Annual Meeting of the Association for
               Computational Linguistics (ACL)",
  pages     = "1621--1630",
  year      =  2016,
  doi       = "10.18653/v1/P16-1153"
}

@ARTICLE{Choset2001-ck,
  title   = "{Topological simultaneous localization and mapping (SLAM): Toward
             exact localization without explicit localization}",
  author  = "Choset, Howie and Nagatani, Keiji",
  journal = "IEEE Trans. Rob. Autom.",
  volume  =  17,
  number  =  2,
  pages   = "125--137",
  month   =  apr,
  year    =  2001,
  doi     = "10.1109/70.928558",
  issn    = "1042-296X"
}

@ARTICLE{Kuipers1991-dt,
  title     = "{A robot exploration and mapping strategy based on a semantic
               hierarchy of spatial representations}",
  author    = "Kuipers, Benjamin and Byun, Yung Tai",
  journal   = "Rob. Auton. Syst.",
  publisher = "North-Holland",
  volume    =  8,
  number    = "1-2",
  pages     = "47--63",
  month     =  nov,
  year      =  1991,
  doi       = "10.1016/0921-8890(91)90014-C",
  issn      = "0921-8890"
}

@ARTICLE{Savinov2018-pg,
  title     = "{Semi-parametric Topological Memory for Navigation}",
  author    = "Savinov, Nikolay and Dosovitskiy, Alexey and Koltun, Vladlen",
  journal   = "6th International Conference on Learning Representations, ICLR
               2018 - Conference Track Proceedings",
  publisher = "International Conference on Learning Representations, ICLR",
  month     =  mar,
  year      =  2018,
  eprint    = "1803.00653"
}

@INPROCEEDINGS{Rublee2011-orb,
  title     = "{ORB: An efficient alternative to SIFT or SURF}",
  author    = "Rublee, Ethan and Rabaud, Vincent and Konolige, Kurt and Bradski,
               Gary",
  booktitle = "2011 International Conference on Computer Vision (ICCV)",
  pages     = "2564--2571",
  year      =  2011,
  doi       = "10.1109/ICCV.2011.6126544"
}

@ARTICLE{Kabsch1976-rt,
  title   = "{A solution for the best rotation to relate two sets of vectors}",
  author  = "Kabsch, Wolfgang",
  journal = "Acta Crystallographica Section A",
  volume  = "32",
  number  = "5",
  pages   = "922--923",
  year    =  1976,
  doi     = "10.1107/S0567739476001873"
}

@ARTICLE{Umeyama1991-ls,
  title   = "{Least-squares estimation of transformation parameters between two
             point patterns}",
  author  = "Umeyama, Shinji",
  journal = "IEEE Transactions on Pattern Analysis and Machine Intelligence",
  volume  = "13",
  number  = "4",
  pages   = "376--380",
  year    =  1991,
  doi     = "10.1109/34.88573"
}

@INPROCEEDINGS{Zhao2021-vo,
  title     = "{The Surprising Effectiveness of Visual Odometry Techniques for
               Embodied PointGoal Navigation}",
  author    = "Zhao, Xiaoming and Agrawal, Harsh and Batra, Dhruv and Schwing,
               Alexander G",
  booktitle = "2021 IEEE/CVF International Conference on Computer Vision (ICCV)",
  pages     = "16107--16116",
  year      =  2021,
  doi       = "10.1109/ICCV48922.2021.01582"
}

@INCOLLECTION{Mazuran2017-rt,
  title     = "{Relative Topometric Localization in Globally Inconsistent Maps}",
  author    = "Mazuran, Mladen and Boniardi, Federico and Burgard, Wolfram and
               Tipaldi, Gian Diego",
  booktitle = "Robotics Research (ISRR 2015)",
  series    = "Springer Proceedings in Advanced Robotics",
  publisher = "Springer",
  pages     = "435--451",
  year      =  2018,
  doi       = "10.1007/978-3-319-60916-4_25"
}

@ARTICLE{Trullier1997,
  title   = "{Biologically based artificial navigation systems: Review and prospects}",
  author  = "Trullier, Olivier and Wiener, Sidney I and Berthoz, Alain and Meyer,
             Jean-Arcady",
  journal = "Prog. Neurobiol.",
  volume  = "51",
  number  = "5",
  pages   = "483--544",
  month   =  apr,
  year    =  1997,
  doi     = "10.1016/S0301-0082(96)00060-3",
  issn    = "0301-0082"
}

@ARTICLE{FranzMallot2000,
  title   = "{Biomimetic robot navigation}",
  author  = "Franz, Matthias O and Mallot, Hanspeter A",
  journal = "Robot. Auton. Syst.",
  volume  = "30",
  number  = "1--2",
  pages   = "133--153",
  month   =  jan,
  year    =  2000,
  doi     = "10.1016/S0921-8890(99)00069-X",
  issn    = "0921-8890"
}

\end{document}